\documentclass[final,5p,times,twocolumn]{elsarticle}

\usepackage{amssymb}
\usepackage{subfigure}
\usepackage{multicol}
\usepackage{multirow}

\usepackage{amsmath}
\usepackage{caption}

\usepackage[table]{xcolor}
\definecolor{mygray}{gray}{.9}

\usepackage[pagebackref=true,breaklinks=true,colorlinks,citecolor=citecolor,bookmarks=false]{hyperref}

\newlength\savewidth

\usepackage{booktabs}

\makeatletter
\newcommand{\thickhline}{%
	\noalign {\ifnum 0=`}\fi \hrule height 1pt
	\futurelet \reserved@a \@xhline
}
\makeatother

\usepackage{newfloat}
\usepackage{listings}
\journal{Neurocomputing}

\begin{document}

\begin{frontmatter}



\title{UV R-CNN: Stable and Efficient Dense Human Pose Estimation}


\author[1]{Wenhe Jia}
\ead{jiawh@bupt.edu.cn}

\author[1]{Yilin Zhou}
\ead{ylzhou@bupt.edu.cn}

\author[2]{Xuhan Zhu}
\ead{zhuxuhan21@mails.ucas.ac.cn}

\author[1]{Mengjie Hu}
\ead{mengjie.hu@bupt.edu.cn}

\author[1]{Chun Liu}
\ead{chun.liu@bupt.edu.cn}

\author[1]{Qing Song\corref{mycorrespondingauthor}}
\cortext[mycorrespondingauthor]{Corresponding author}
\ead{priv@bupt.edu.cn}

\affiliation[1]{
	organization={Beijing University of Posts and Telecommunications, Artificial Intelligence Academy},%Department and Organization
	addressline={10th xitucheng road, Haidian District}, 
	city={Beijing},
	postcode={100086}, 
	state={Beijing},
	country={China}
}

\affiliation[2]{
	organization={Chinese Academy of Sciences, Institute of Computing Techonolgy},%Department and Organization
	addressline={}, 
	city={Beijing},
	postcode={100086}, 
	state={Beijing},
	country={China}
}

%%%%%%%%%%% Abstract %%%%%%%%%%%
\begin{abstract}	
Dense pose estimation is a dense 3D prediction task for instance-level human analysis, aiming to map human pixels from an RGB image to a 3D surface of the human body. Due to a large amount of surface point regression, the training process appears to be easy to collapse compared to other region-based human instance analyzing tasks. By analyzing the loss formulation of the existing dense pose estimation model, we introduce a novel point regression loss function, named \emph{Dense Points} loss to stable the training progress, and a new balanced loss weighting strategy to handle the multi-task losses. With the above novelties, we propose a brand new architecture, named UV R-CNN. Without auxiliary supervision and external knowledge from other tasks, UV R-CNN can handle many complicated issues in dense pose model training progress, achieving 65.0\% $AP_{gps}$ and 66.1\% $AP_{gpsm}$ on the DensePose-COCO validation subset with ResNet-50-FPN feature extractor, competitive among the state-of-the-art dense human pose estimation methods.
\end{abstract}


%%%%%%%%%%% Highlights %%%%%%%%%%%
%\item We introduce a novel surface point regression loss function for dense pose estimation, named \emph{Dense Points} loss, which can constraint the loss and gradient of U/V point regression at the begining of training, making training  pregress stable and large initial learning rate applicable.

%\item To handle the multi-task loss in training dense pose models, we propose a balanced loss weighting strategy that enables the model to determine how much each task contributes to the optimization. The proposed weighting strategy brings significant performance improvement.

%\item Without external knowledge and auxiliary supervision from other human instances analyzing tasks and data sources, the proposed architecture achieves 65.0\% $AP_{gps}$ and 66.1\% $AP_{gpsm}$ on the DensePose-COCO validation subset with ResNet-50-FPN feature extractor, competitive among the state-of-the-art dense human pose estimation methods.

\begin{keyword}
Dense pose estimation \sep Model  optimizing \sep Balanced loss weights



\end{keyword}

\end{frontmatter}


%%%%%%%%%%% sec 1 %%%%%%%%%%%
\section{Introduction}
\label{sec:intruduction}
\begin{figure*}[!t]
	\centering
	\subfigure[Image.]{
		\label{fig:Image}
		\begin{minipage}[t]{0.23\linewidth}
			\centering
			\includegraphics[width=1.0\linewidth]{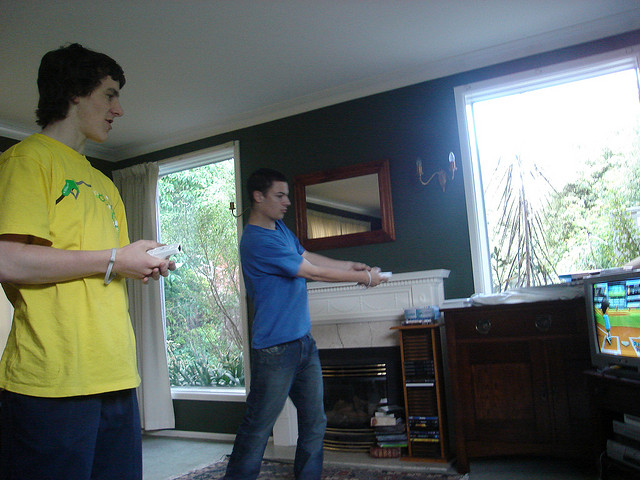}
		\end{minipage}
	}
	\subfigure[Parsing.]{
	\label{fig:Image}
	\begin{minipage}[t]{0.23\linewidth}
		\centering
		\includegraphics[width=1.0\linewidth]{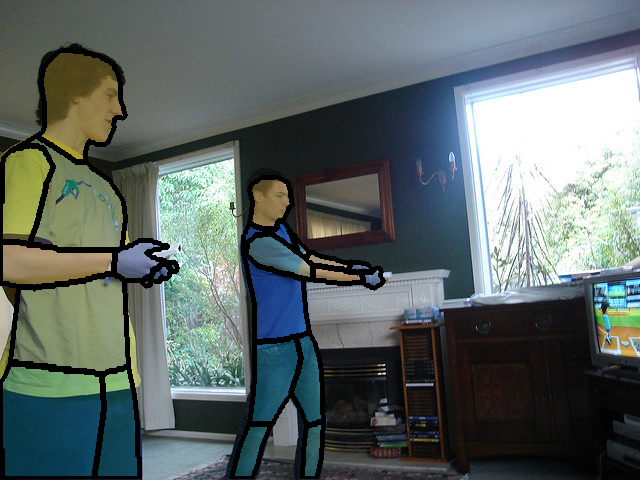}
	\end{minipage}
	}
	\subfigure[U coordinates.]{
	\label{fig:U-coordinates}
	\begin{minipage}[t]{0.23\linewidth}
		\centering
		\includegraphics[width=1.0\linewidth]{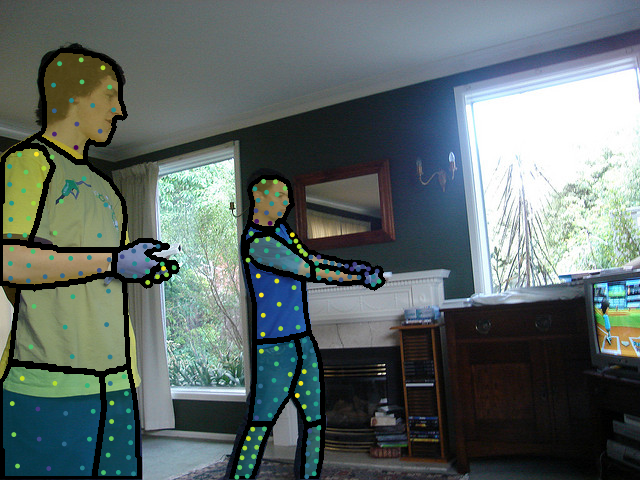}
	\end{minipage}
	}
	\subfigure[V coordinates.]{
	\label{fig:V-coordinates}
	\begin{minipage}[t]{0.23\linewidth}
		\centering
		\includegraphics[width=1.0\linewidth]{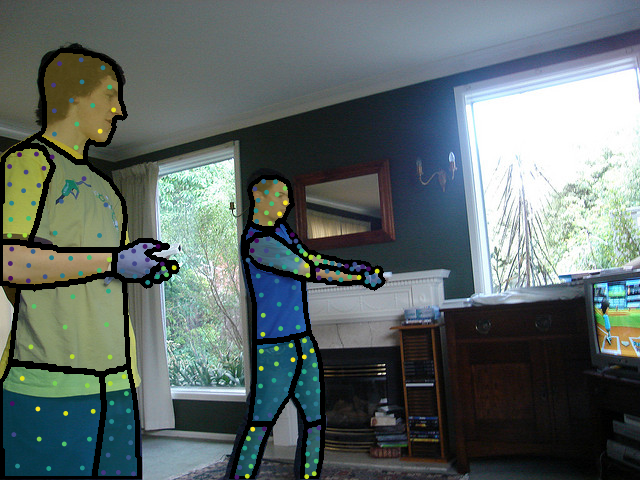}
	\end{minipage}
	}
	\captionsetup{font={normalsize}, justification=raggedright}
	\caption{Visualization of the DensePose-COCO dataset annotations. (a) The original image. (b) The human-part segmentation mask. (c) The human-part-specific U coordinates. (d) The human-part-specific V coordinates.}
	\label{Densepose-Data}
\end{figure*}

Human instance analyzing technique is an essential component of artificial intelligence applications in the real world, such as human pose estimation \cite{convposemachine, hourglass, openpose, baptista2020towards, lightweight3dpose, 3dposevideo}, human part segmentation\cite{pgn, li2018multi, lip, zhao2018understanding} and human-object interactions\cite{gao2018ican, detectinginteractions, relationindetection}. Predicting human details at the instance level, these tasks arrive at the same end by different means. Inspired by \cite{pgn} on learning correspondences between RGB images and 3D surfaces, Guler \textit{et al.} proposed a new and challenging task, Dense Human Pose Estimation\cite{guler2018densepose}, establishing a correspondence between the 2D human image pixel and the 3D human body surface. Focusing on humans as the primary object with a brand new DensePose-COCO dataset, dense human pose estimation requires more human body details at the instance level, such as the visibility of different human body surfaces under a variety of human poses, as shown in Figure.\ref{Densepose-Data}. Successful development of convolutional neural networks\cite{rcnn,maskrcnn,imagenet,fcn,inception,gn}has made great progress in instance-level human analysis.  Several outstanding works\cite{guler2018densepose,parsingrcnn} for dense human pose estimation followed a top-down method, demonstrated by Mask R-CNN\cite{maskrcnn}, which first detects all objects in an image and then predicts a class-aware segmentation mask in each instance. Mask R-CNN established a brief and efficient pipeline for instance analyzing, achieved great success in object detection, instance segmentation, human pose estimation, \emph{etc}. Motivated by Mask R-CNN, \cite{guler2018densepose} adopted the two-stage pipeline for dense human pose estimation and proposed a fundamental architecture named DensePose R-CNN. Particularly, they first detected all human instances of an image and then solved the single-person densepose estimation problem.

However, there are still several deficiencies in achieving better performance for existing dense human pose estimation methods in the wild.

First, imbalance loss arrangement for multi-task learning. Different from other instance-level analyzing task solved by region-based solution, Faster R-CNN for object detection as an example, densepose estimation need to compute additional four losses in the UV-branch. Proper arrangement for weights of different losses in the previous works is missing, they all follow a hand-crafted and experience-based method, requiring much time to find an appropriate set of hyperparameters.

Second, low training efficiency. Due to the relatively large number of human body surface points to be regressed, the learning rate during the training phase needs to be minor, enabling the predicted UV coordinates to progressively and stably approach the ground-truth labels. \emph{Smooth-L1} loss\cite{guler2018densepose} was widely adopted to perform the regression of UV coordinates, with large loss value and unstable gradients for backpropagation.

This paper explores the above dense human pose estimation problems. Through in-depth analysis of loss formulation and training strategy, we develop novel improvements to stabilize and speed up the training progress. Adopting elaborate design of the dense pose branch, all the counterparts formed the proposed UV R-CNN architecture, achieving competitive results on DensePose-COCO benchmark.

Our goal is to optimize the unified pipeline for dense human pose estimation without any auxiliary supervision and external knowledge from other aspects, as well as boost the model's ability to perform detailed human analysis in the wild. The contributions of this paper are summarized as follows:
\begin{itemize}
	\item We introduce a novel \emph{Dense-Point} loss to stable the training process. At the beginning of training, the proposed loss function can generate smaller losses and gradients when the predicted UV coordinates are relatively far from the ground truth. A more stable and efficient training process is achieved with more significant learning rates.

	\item Novel balanced multi-task loss weighting strategy. With only a few parameters to be manually designed, a balanced weighting strategy automatically weights losses for training dense pose network, significantly improving the previous pipelines.

	\item Third, inspiring results. We improved the dense pose estimation task with the proposed UV R-CNN over the previous works. On the DensePose-COCO validation subset, UV R-CNN achieves 65.6\% mAP with ResNet-50 as backbone, surpassing the DensePose R-CNN and Parsing R-CNN by 14.5\% mAP and 6.7\% mAP, respectively, competitive with the state-of-the-art results without any auxiliary supervision and external knowledge.
\end{itemize}

\begin{figure*}[!t]
	\centering
	\includegraphics[scale=0.15]{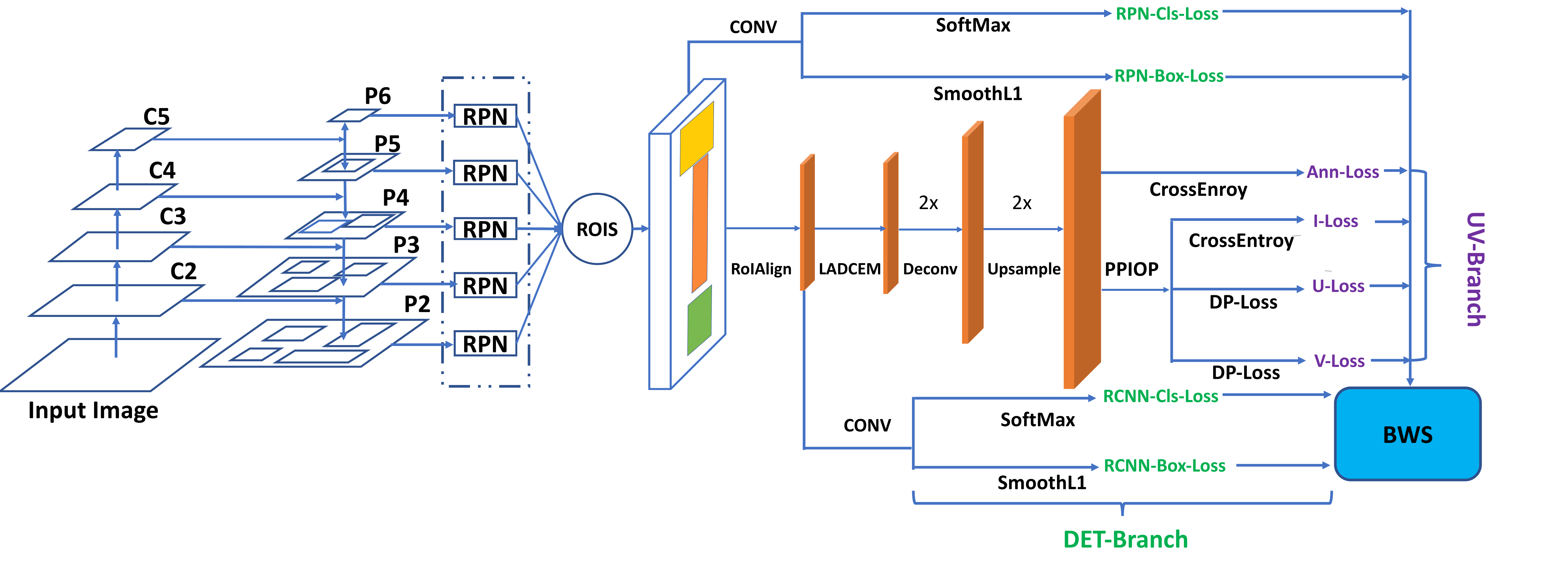} 
	\captionsetup{font={normalsize}, justification=raggedright}
	\caption{The pipeline of our proposed UV R-CNN follows the two-stage region-based instance analyzing method. We adopt the FPN with backbone and RoIAlign operation. UV-Branch is used for dense pose estimation. Det-Branch is used for human instance detection. PPIOP means Pool Points Interpolation Operation proposed in \cite{guler2018densepose}.}
	\label{fig:pipline}
\end{figure*}

\section{Related Work}
\label{sec:relatedwork}
\subsection{Human analysis}
\label{sec:humanbodyanalysis}
In recent years, human body analysis has achieved remarkable developments, such as human detection \cite{Heirrcnn,Cpmrcnn}, human body key points detection, human body part segmentation\cite{parsingrcnn,Pdrnet,Rprcnn}, \emph{etc}. Thanks to the rapid development of object detection, human detection has also made remarkable progress. The anchor-based detectors of two-stage\cite{fastrcnn,fasterrcnn,Airnet,cascadercnn,librarcnn} and one-stage\cite{yolo,ssd,retinanet} pipelines dominates the first five years of the object detection field with densely predefined anchor boxes. Then the anchor-free schemes appear in recent years, considering both speed and accuracy \cite{fcos}. With the prosperity of Vision Transformer \cite{ViT} in computer vision, object detector performance has been pushed to a relatively higher level than ever before\cite{detr,deformabledetr}. Human body segmentation is also an essential task of human body analysis, developing with several widely used benchmarks\cite{MHP,pgn}. Ke Gong \emph{et al.} propose Nested for performing semantic saliency prediction, instance-agnostic parsing, and instance-aware clustering\cite{pgn}. For human body keypoint detection,  The two-word proposition studied in recent years mainly predicts 2D key points\cite{convposemachine,openpose,hourglass} and 3D key points\cite{baptista2020towards,lightweight3dpose,3dposevideo} for human body keypoint detection. The prediction of 3D key points is
somewhat similar to the dense pose estimation task, but the main difference is that the latter prediction input images are 2D images, not 3D images.

\subsection{Dense Pose Estimation}
\label{sec:densepose}
DensePose R-CNN\cite{guler2018densepose} was the first to solve the problem of building dense correspondences between a single 2D image and a 3D human surface. With the proposed DensePose-COCO dataset, Guler \emph{et al.}. Divided the human body into 14 semantic parts and divided each into 1-2 patches for estimating the part-specific UV coordinates. Finally, they collected about 196 UV coordinates for each person. Subsequently, Yang \emph{et al.} followed the two-stage pipeline to estimate the UV coordinates for each human body pixel, extracting the RoI feature with more human instance details and strengthening geometric and context encoding of different human body parts\cite{parsingrcnn}, Parsing R-CNN largely facilitates the performance on dense pose estimation. Recently, \cite{amanet} and \cite{ktn} use multi-task learning strategy and knowledge transfer method to enhance the model's ability to map image pixels to the human body surface. \cite{amanet} utilizes a adaptive multi-path aggregation module to capture rich objects  information from different feature resolution, then adaptively leverages them in multiple human analysis task(\emph{e.g}., instance segmentation, body part segmentation, and UV estimation). \cite{ktn} introduces a knowledge transfer machine to transfer the model weights learned from keypoint annotation into body surface classifcation sub-model, which serves as an auxiliary commonsense of human body.

\subsection{Multi-task Learning}
\label{sec:multitrasklearning}
Multi-task learning refers to multiple tasks simultaneously, such as learning classification and regression tasks. In computer vision, multi-task learning has been used for learning many tasks such as image classification in multiple domains\cite{rebuffi2017learning}, pose estimation and action recognition\cite{gkioxari2014r}, and dense prediction of depth, surface normals, and semantic classes \cite{misra2016cross,eigen2015predicting}. An essential problem in multi-task learning is balancing the weights between different tasks. Several previous works argue that different amounts of sharing and weighting tend to work best for different tasks\cite{misra2016cross,cipolla2018multi}. Dynamic Task Prioritisation\cite{guo2018dynamic} encourages prioritization of complex tasks directly using performance metrics such as accuracy and precision. \cite{liu2019end} introduce a soft-attention module to obtain each task's valuable features for calculation from the leading network and finally achieve the effect of multi-task calculation. However, they can’t adjust the weight between each task loss in an automatic and balanced manner. This paper uses an automatic strategy to balance losses of different tasks with little hyperparameters.

\section{UV R-CNN}
\label{sec:uvrcnn}
As shown in Figure.~\ref{fig:pipline}, the overall architecture of UV R-CNN follows the region-based approach proposed by DensePose R-CNN\cite{guler2018densepose} and improved by Parsing R-CNN\cite{parsingrcnn}, utilizing an end-to-end trained CNN model consisting of backbone, region proposal network(RPN) and several branches as task-specific networks. In this section, we will introduce the proposed design of the proposed UV R-CNN in detail.

\begin{figure*}[!t]
	\centering
	\subfigure[AP/LR]{
		\label{fig:AP/LR}
		\begin{minipage}[t]{0.48\linewidth}
			\centering
			\includegraphics[width=1.0\linewidth]{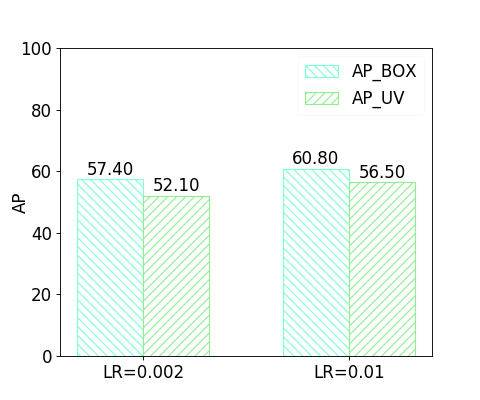}
		\end{minipage}
	}
	\subfigure[DP-Loss.]{
		\label{fig:DPvsSmooth} 
		\begin{minipage}[t]{0.4\linewidth}
			\centering
			\includegraphics[width=1.0\linewidth]{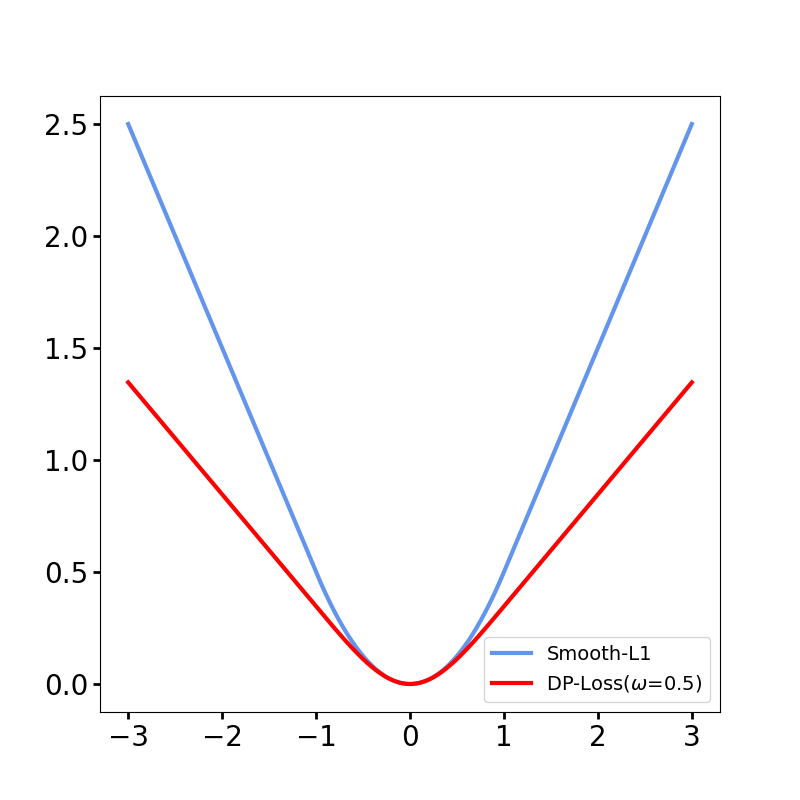}
		\end{minipage}
	}
	\captionsetup{font={normalsize}, justification=raggedright}
	\caption{(a) Histogram of the human category AP of box and uv in DensePose-COCO validation dataset, other hyperparameters are the same as in \cite{guler2018densepose}. (b) Plots of the DP-Loss and Smooth L1 loss functions. LR: Learning Rate.}
	\label{fig:DP-Loss}
\end{figure*}

\subsection{Dense Points Loss}
\label{sec:densepointsloss}
At the beginning of the DensePose R-CNN training progress, there are always some lousy UV points regressed whose coordinates are far from the ground truth. These outlied UV points bring with the non-negligible burden to the UV regression driven by \emph{Smooth-L1} loss, immense loss and gradients leading to unstable training at the beginning. To handle this, DensePose R-CNN and Parsing R-CNN tried to set a small learning rate compared to training the object detector alone as a compromise solution. This will severely damage the model's performance for object detection, as shown in Figure.~\ref{fig:AP/LR}. 

In the region-based human analyzing methods, the object detection task is the premise of other tasks. An inaccurate bounding box may exclude annotated UV points belonging to the human surface. To stabilize the initial training stage, we propose a modified \emph{Smooth-L1} loss function, named \emph{Dense Points} loss(DP-Loss).

\begin{equation}
	\mathcal{L}_{DP}(x)=
	\left\{
	\begin{aligned}
		\omega \ln{(1+x^2)} &  & {if \mid x \mid < 1, 0< \omega <= \frac{1}{2}} \\
		\omega [\mid x \mid +(\ln{2}-1)] &  & {otherwise, 0< \omega <= \frac{1}{2}}
	\end{aligned}
	\label{eq:dploss}
	\right.
\end{equation}

The computation of Dense Point loss is shown in Eq~\ref{eq:dploss}, the gradients of \emph{Dense Point} loss is calculated as Eq~\ref{eq:graddp}:

\begin{equation}
	Grad_{DP}(x)=
	\left\{
	\begin{aligned}
		\frac{\omega \mid x \mid}{1+x^2} &  & {if \mid x \mid < 1, 0< \omega <= \frac{1}{2}} \\
		\omega &  & {otherwise, 0< \omega <= \frac{1}{2}}
	\end{aligned}
	\label{eq:graddp}
	\right.
\end{equation}

For comparison, the gradient of \emph{Smooth L1} loss is as Eq~\ref{eq:gradsmooth}:

\begin{equation}
	Grad_{Smooth-L1}(x)=
	\left\{
	\begin{aligned}
		\mid x \mid &  & {if \mid x \mid < 1} \\
		1 &  & {otherwise}
	\end{aligned}
	\label{eq:gradsmooth}
	\right.
\end{equation}

Described by Figure.~\ref{fig:DPvsSmooth} and Eq.~\ref{eq:dploss}, \ref{eq:graddp}, \ref{eq:gradsmooth}, comparing with \emph{Smooth L1} loss function, Dense Point loss has relatively more minor loss values and gradients under the same network input and hyperparameter $\omega$, enabling it to be more robust to outlied UV points. A more significant learning rate can be adopted to accelerate the model training, ensuring good performance of both object detection and UV coordinates regression.

\subsection{Balanced Weighting Strategy}
\label{sec:bws}
There are eight losses arranged in the end-to-end pipeline of dense pose estimator, such as DensePose R-CNN and Parsing R-CNN, total loss constructed as Eq~\ref{eq:losscombination} is decoupled into two parts: detection losses and dense pose losses, dense pose losses are multiplied by decimal \emph{K} to be combined with RPN and RCNN detection losses, the dense pose losses consisting of human part classification loss, U/V coordinates losses and UV index loss, the minimization of these losses can be seen as a multi-task learning progress.

\begin{equation}
	\begin{aligned}
		\mathcal{L}_{all} &= \mathcal{L}_{det} + K\mathcal{L}_{uv}\\
		&= \mathcal{L}_{rpn\_cls} + \mathcal{L}_{rpn\_reg} + \mathcal{L}_{rcnn\_cls} + \mathcal{L}_{rcnn\_reg} \\
		&+ K(\mathcal{L}_{Ann} + \mathcal{L}_{I} + \mathcal{L}_{U} + \mathcal{L}_{V})
	\end{aligned}
	\label{eq:losscombination}
\end{equation} 

For multi-task learning, all relevant tasks should be equally treated, the contribution of each task should be considered in a balanced way in the learning process rather than allowing one task to be dominant and thus affecting the optimization of other tasks. In the region-based dense pose model, the loss function of each task has different calculation methods and normalization strategies. The combined loss of tasks can degrade the accuracy of one or all of them. Especially due to a large number of UV coordinates to be regressed, the loss value of the UV branch is relatively larger compared to the other region-based human instance analysing tasks, \emph{e.g.}, human keypoint detection task, larger gradients of UV branch will disturb the learning of other model components.

% table 1
\begin{table}[t]
	\centering
	\captionsetup{font={normalsize}, justification=raggedright}
	\caption{The initial weights corresponding to each loss in DensePose-RCNN and Parsing R-CNN.}
	\label{table:lossweight}
	\setlength\tabcolsep{3.0mm}
	\renewcommand\arraystretch{1.4}
	\scalebox{1.0}{
		\begin{tabular}{l|l|c|l|l|c}
			\hline\thickhline
			%\cline{0-1}\cline{3-4}
			\rowcolor{mygray}
			\multicolumn{2}{c}{Loss} & Weight & \multicolumn{2}{c}{Loss} & Weight \\
			\cline{1-4}
			\hline
			\hline
			%\cline{1-2}\cline{3-4}
		    & $\mathcal{L}$$_{rpn\_cls}$ & 1.0 & &$ \mathcal{L}$$_{Ann}$ & 2.0 \\
			& $\mathcal{L}$$_{rpn\_reg}$ & 1.0 & & L$_I$ & 0.3 \\
			& $\mathcal{L}$$_{rcnn\_cls}$ & 1.0 & & L$_U$ & 0.1 \\
			\multirow{-4}{*}{$\mathcal{L}$$_{det}$} & $\mathcal{L}$$_{rcnn\_reg}$ & 1.0 & \multirow{-4}{*}{$\mathcal{L}$$_{uv}$}  & L$_V$ & 0.1 \\
			\hline
		\end{tabular}
	}
\end{table}

The scale factors of UV losses are empirical numbers, which need to be manually adapted to achieve good performance. Manually adjusted weights of different losses in DensePose-RCNN are listed in Table~\ref{table:lossweight}. For easy implementation, researchers used to perform grid search on these scale factors to find the best set of hyperparameters, requiring tedious experiment time cost. When the grid density increases, the search space will boost with the increase of geometric series, followed by a significant decline in experimental efficiency. An ideal solution is to adopt an automatic manner to learn the corresponding loss weights. To address this, we propose a balanced weighting strategy that enables the model to determine how much each task contributes to the optimization. 

\begin{equation}
	\mathcal{L} = \sum\limits_{i=1}^{n}\mathcal{L}_{i}\times \frac{exp^{-\ \mathcal{L}_{i}}} {\sum\limits_{i=1}^{n}exp^{-\ \mathcal{L}_{i}}}
	\label{eq:totalloss}
\end{equation}

% table 2
\begin{table*}[!h]
	\centering
	\captionsetup{font={normalsize}, justification=raggedright}
	\caption{The initial weights corresponding to each loss in DensePose-RCNN and Parsing R-CNN.}
	\setlength\tabcolsep{6.0mm}
	\renewcommand\arraystretch{1.4}
	\scalebox{1.0}{
		\begin{tabular}{c|c|c|c|c|c}
			\hline\thickhline
			\rowcolor{mygray}
			Method & $\omega$ & LR & init weight $\mathcal{L}_{U,V}$ & $\qquad$AP$_{gps}$$\qquad$ & $\qquad$AP$_{gps^m}$$\qquad$ \\
			\hline
			baseline & - &  0.002  & 0.1 & 51.0 & - \\
			DP-Loss &0.5  &0.002 & 0.1 & 52.4   &  55.4 \\
			DP-Loss &0.25  &0.002 & 0.1 & \textbf{53.1} & \textbf{56.0} \\
			\hline
			baseline & - &  0.01  & 0.05 & 56.5 & 59.0 \\
			baseline & -&0.02  &0.05& $Error^{\dagger}$ & $Error^{\dagger}$ \\
			DP-Loss & 0.5&0.02 & 0.05 & 57.9  & 60.1  \\
			DP-Loss & 0.25&0.02 & 0.05 &\textbf{58.5} & \textbf{60.5}\\
			\hline
		\end{tabular}
	}
	\label{table:dploss}
\end{table*}

As shown in Eq.~\ref{eq:totalloss}, for each part of two loss components, the proposed balanced weighting strategy(BWS) first takes the negative power of each loss sub-component, then calculates the percentage of each sub-components as the loss weight, the usage of negative exponential can transform the value of the losses to a normalized range [0, 1], the sub-losses are then multiplied by loss weights and combined in a weighted sum manner. Thus the training progress can automatically adjusts the loss weights according to the contribution of each loss component at a single iteration. The task with a relatively more significant loss value at the current iteration will be less considered in the next iteration. The total loss of dense pose model during training can be calculated as Eq.~\ref{eq:bwsloss}. Here, we keep the weight factor between human detection loss and dense pose UV estimation loss as a hyperparameter.

\begin{equation}
	\begin{aligned}
		\mathcal{L}_{BWS} 
		 = \sum\limits_{i=1}^{n}\mathcal{L}_{det_{i}}\times \frac{exp^{-\ \mathcal{L}_{det_{i}}}} {\sum\limits_{i=1}^{n}exp^{-\ \mathcal{L}_{det_{i}}}}\\
		+ K\times \sum\limits_{i=1}^{n}\mathcal{L}_{UV_{i}}\times \frac{exp^{-\ \mathcal{L}_{UV_{i}}}} {\sum\limits_{i=1}^{n}exp^{-\ \mathcal{L}_{UV_{i}}}}
	\end{aligned}
	\label{eq:bwsloss}
\end{equation}

\subsection{DensePose Network Improvements}
\label{sec:networkimpr}
Human instance and its different parts usually vary in vision scale greatly, which requires the feature maps to capture multi-scale information, and the geometrically related human parts should be jointly considered. The feature of each human body part should contain the whole body layout. A simple stack of convolution layers with limited spatial attention is not enough. Also, the extracted 14$\times$14 RoI features from image feature maps of different resolutions suffer too much loss of detailed information of human instance.

To address these drawbacks existing in the simple branch network, Parsing R-CNN\cite{parsingrcnn} proposed Proposals Separation Sampling(\emph{PSS}), Enlarging RoI Resolution(\emph{ERR}) and  Geometric and Context Encoding(\emph{GCE}) module to improve the performance of human parsing network, respectively. In PSS, the bbox branch still adopts the scale assign strategy on the feature pyramid according to FPN. However, the RoIPool/RoIAlign operation of the parsing branch is only performed on the finest scale feature map of P2. Accompanied by ERR(32$\times$32 out the resolution of RoIAlign), the parsing sub-network can preserve more human body details. Then, GCE consists of Non-Local\cite{nonlocal} and ASPP\cite{deeplabv3} to capture geometric human body correlation and multi-scale information, which also benefits the dense human pose estimation. Later, Zhu \emph{et al.}\cite{zhu2021joint} introduced an improved version of the GCE module to capture human geometric and context information in a more enhancing and cost-effective manner, named LADCEM.

This work adopts the previously proposed PSS, ERR, and LADCEM improvements to the dense pose network.

\section{Experiments}
\label{sec:exp}
In this section, we compare the performance of proposed UV R-CNN with state-of-the-art methods, we also perform ablation studies to explore and verify the effectiveness of the proposed components. 

\subsection{Implementation Details}
\label{sec:impld}

\textbf{Dataset}. We conduct extensive experiments on DensePose-COCO\cite{guler2018densepose} dataset, which is a large-scale human body analysis dataset with images and surfaces manually annotated on 50K COCO\cite{coco} images. Twenty-four types of crude body parts are present as points, and 14 types of fine body parts are present as masks. In addition, there is UV annotation which is most 196 points and ranges from (0,1].

\textbf{Metircs}. Follwing\cite{guler2018densepose}, we adopt the Average Precision (AP) at a number of geodesic point similarty (GPS) thresholds ranging from 0.5 to 0.95 as the evaluation metric, its formulation is as following:

\begin{equation}
	GPS_{j} = \frac{1}{\vert P_{j} \vert }\sum\limits_{p\in P_{j}}exp^(\frac{-g(i_{p}, \hat{i}_{p})^2}{2\kappa^2})
	\label{eq:gpsj}
\end{equation}

Specially, the organizer develope a new metric called masked geodesic point similarity in the 2019 COCO DensePose-Chanllage, calculated as:

\begin{equation}
	GPS^m = 
	\sqrt{GPS\cdot \mathcal{I}},    \mathcal{I} = \frac{\mathcal{M}\cap \mathcal{\hat M}}{\mathcal{M}\cup \mathcal{\hat M}}
	\label{eq:gpsm}
\end{equation}where \emph{GPS} is the geodesic point similarity metric value and and $\mathcal{I}$ is  the intersection over union between the ground truth ($\mathcal{M}$) and the predicted ($\mathcal{\hat M}$) foreground masks. We will use all the two different metrics for reference by subsequent researchers.

\textbf{Baseline}. We design a relatively simple baseline model to demonstrate the effectiveness of our proposed improvements for DensePose estimation. Firstly, a ResNet-50\cite{resnet} with Feature Pyramid Network\cite{fpn} is used as a feature extractor. Then, an object detection branch with two fully connected layers is adopted to predict classification scores and bounding box coordinates for each RoI. Last, a stack of eight 3×3 512-d convolutional layers followed by a deconvolution\cite{deconv} layer with stride 2 and a 2× bilinear up-sampling layer form the dense pose branch. 

% table 3
\begin{table*}[!t]
	\centering
	\captionsetup{font={normalsize}, justification=raggedright}
	\caption{Ablation study on Balanced Weighting Strategy. $Error^{\dagger}$ means if using the LR (learning rate) as 0.04, the training process will crash.}
	\setlength\tabcolsep{5.0mm}
	\renewcommand\arraystretch{1.4}
	\scalebox{1.0}{
		\begin{tabular}{c|c|c|c|c|c|c}
			\hline\thickhline
			\rowcolor{mygray}
			Method & DP-Loss & LR & init weight $\mathcal{L}_{U,V}$ & K & $\qquad$AP$_{gps}$$\qquad$ & $\qquad$AP$_{gps^m}$$\qquad$ \\
			\hline
			\hline
			baseline & \checkmark & 0.002  & 0.1 & 1.0 & 53.1 & 56.0  \\
			BWS & \checkmark & 0.002 & 0.1 & 1.0 & 51.7  &  53.9 \\
			\hline
			baseline & \checkmark &  0.02  & 0.05 & 1.0 & \textbf{58.5} & \textbf{60.6}  \\
			BWS & \checkmark & 0.02  & 0.05 & 1.0 & 56.9 &  59.8 \\
			baseline & \checkmark & 0.04  & 0.05 & 1.0 & $Error^{\dagger}$ & $Error^{\dagger}$ \\
			BWS & \checkmark & 0.04 & 0.05 & 1.0 & 57.1  &  59.6 \\
			BWS & \checkmark & 0.06 & 0.05 & 1.0 & 57.6 & 59.9 \\
			\hline
			BWS & \checkmark & 0.04  & 0.05 & 2.0 & 58.3 &  60.4 \\
			BWS & \checkmark & 0.06 & 0.05 & 2.0 & \textbf{59.1} & \textbf{61.2} \\
			\hline
		\end{tabular}
	}
	\label{table:BWS1}
\end{table*}

%table 4
\begin{table*}[!h]
	\centering
	\captionsetup{font={normalsize}, justification=raggedright}
	\caption{Dense pose estimation results on DensePose-COCO validation dataset. We adopt ResNet50-FPN as backbone. The baseline is DensePose-RCNN.}
	\setlength\tabcolsep{9.0mm}
	\renewcommand\arraystretch{1.4}
	\scalebox{1.0}{
		\begin{tabular}{cccc|cc}
			\hline\thickhline
			\rowcolor{mygray}
			Baseline & DP-Loss & Impr & BWS  & AP$_{gps}$ & AP$_{gps^m}$ \\
			\hline
			\hline
			ResNet-50 & & & & & \\
			\checkmark & & & &  51.0 & - \\
			\checkmark & \checkmark & & &  53.1 &  56.0 \\
			\checkmark & \checkmark & \checkmark & & 57.1 & 59.6  \\
			\checkmark & \checkmark & \checkmark & \checkmark & \textbf{65.0} &  \textbf{66.1} \\
			\hline
			$\triangle$& & & & \textbf{\textbf{+14.0}}& - \\
			\hline
		\end{tabular}
	}
	\label{table:r50ablation}
\end{table*}

Following DensePose R-CNN, the proposed UV-RCNN has experimented with 8 NVIDIA Titan X GPUs. Each mini-batch involves 2 images per GPU; 512 ROIs are sampled from each image. We train all models using image scales randomly sampled from [512, 864] pixels, and single scale inference is performed with the short edge of the image resized to 800 pixels. The training progress has 130k iterations, with the learning rate reducing by 10 at 100k and 120k iterations. The warming-up iterations of the learning rate are 500 with a warming-up factor of 0.1. We set the initial loss weights of $\mathcal{L}_{U, V}$ to 0.1 and the learning rate to 0.002.

\subsection{Ablation Studies on DensePose-COCO}

\textbf{Dense Points Loss.} The ablation studies of Dense Points loss are shown in Table~\ref{table:dploss}. Compared to the baseline model, Dense Points-Loss improves the APgps by 1.4($\omega$=0.5) and 2.1($\omega$=0.25). When increasing the learning rate to 0.02, the training progress driven by Smooth-L1 loss collapses as in \cite{guler2018densepose}, but with Dense Points loss, the model can be trained stably and receive considerable improvements on AP$_{gps}$(58.5) and AP$_{gps^m}$(60.5).
\label{subsec1}

\textbf{Balanced Weighting Strategy.} To verify the effectiveness of BWS, we adopt DP-Loss and set the initial learning rate to 0.02 and weights of $\mathcal{L}_{U, V}$ to 0.05 as our baseline in this subsection. As shown in Table~\ref{table:BWS1}, directly equipping BWS with DensePose R-CNN driven by DP-Loss will cause performance degradation. When increasing the learning rate on the baseline of DP-Loss to 0.04, the training progress collapses, but BWS can make the model successfully trained. When increasing the learning rate to 0.06, BWS brings light improvement to both AP$_{gps}$(59.1) and AP$_{gps^m}$(61.2). We further increase the weight factor K of dense pose losses in Eq~\ref{eq:totalloss} to explore better settings because when K is small, the gradient ratio of the UV branch will reduce. So we adjust K larger, and we can see it will get better precision. All the ablation experiments show that BWS and DP-Loss can make the training progress more efficiently.

% table 5
\begin{table*}[!t]
	\centering
	\captionsetup{font={normalsize}, justification=raggedright}
	\caption{Results of state-of-the-art methods on DensePose-COCO validation subset. All models are trained on DensePose-COCO training subset.}
	\setlength\tabcolsep{12mm}
	\renewcommand\arraystretch{1.4}
	\scalebox{1.0}{
		\begin{tabular}{l|c|c}
			\hline\thickhline
			\rowcolor{mygray}
			BackBone & Method & $\qquad$AP$_{gps}$$\qquad$ \\
			\hline
			\hline
			 & Densepose R-CNN\cite{guler2018densepose} & 51.0 \\
			 & Parsing R-CNN\cite{parsingrcnn} & 58.3 \\
			 & AMA-net\cite{amanet} & 64.1 \\
			 & KTN\cite{ktn} & 66.5 \\
			\multirow{-5}{*}{ResNet-50} & UV-RCNN & \textbf{65.0} \\
			\hline
		 	& Densepose-RCNN & 51.8 \\
			\multirow{-2}{*}{ResNet-101} & UV-RCNN & \textbf{65.5} \\
			\hline
			& Parsing R-CNN & 61.6 \\
			\multirow{-2}{*}{ResNeXt-101} & UV-RCNN & \textbf{61.9} \\
			\hline
		\end{tabular}
	}
	\label{table:SOTA}
\end{table*}

\begin{figure*}[!t]
	\centering
	\begin{minipage}{17cm}
		\includegraphics[width=4cm,height=3cm]{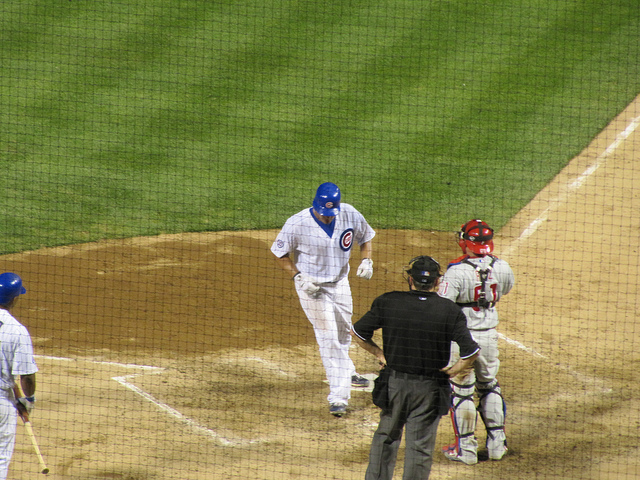}
		\includegraphics[width=4cm,height=3cm]{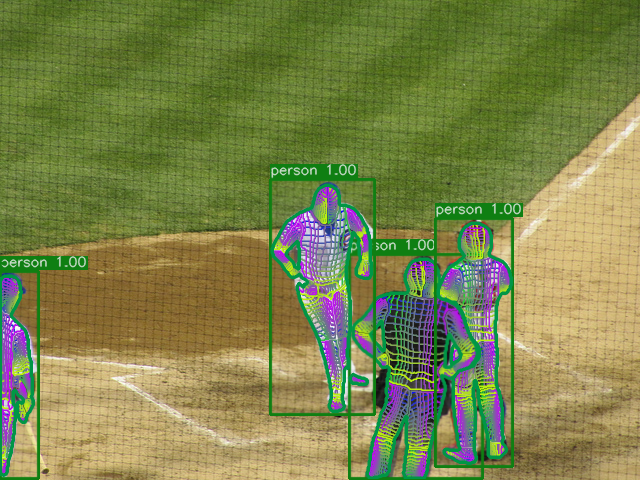}
		\includegraphics[width=4cm,height=3cm]{figs/COCO_val2014_000000000294.jpg}
		\includegraphics[width=4cm,height=3cm]{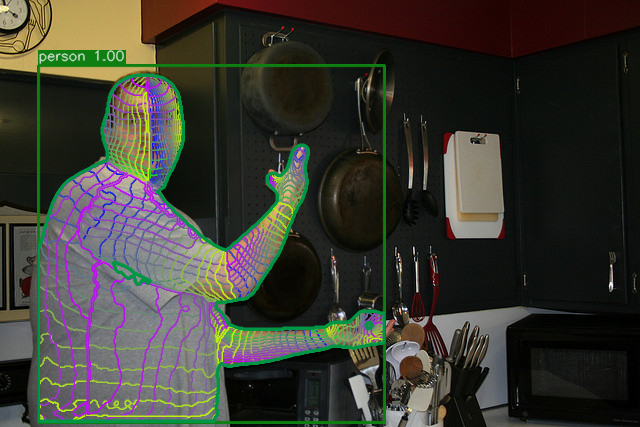}
	\end{minipage}
	
	\begin{minipage}{17cm}
		\includegraphics[width=4cm,height=3cm]{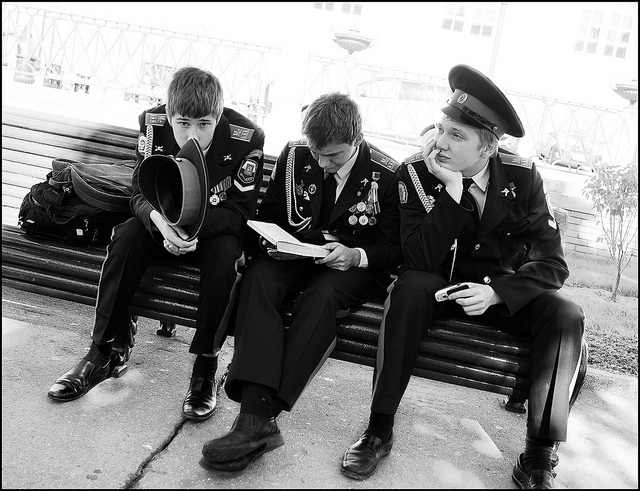}
		\includegraphics[width=4cm,height=3cm]{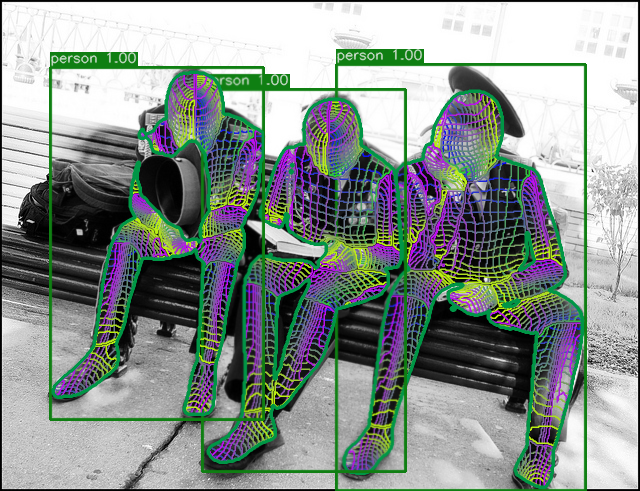}
		\includegraphics[width=4cm,height=3cm]{figs/COCO_val2014_000000000885.jpg}
		\includegraphics[width=4cm,height=3cm]{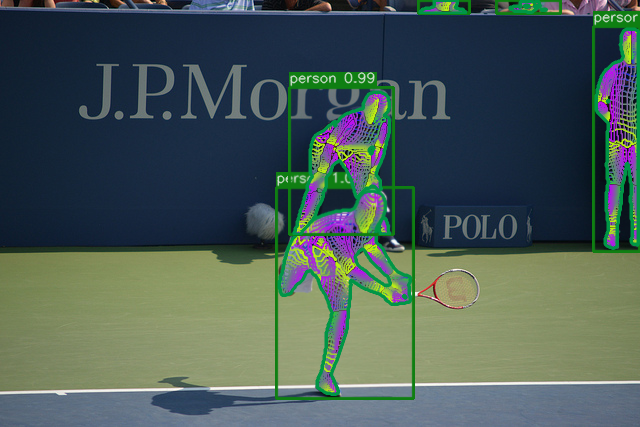}
	\end{minipage}
	
	\begin{minipage}{17cm}
		\includegraphics[width=4cm,height=3cm]{figs/COCO_val2014_000000001268.jpg}
		\includegraphics[width=4cm,height=3cm]{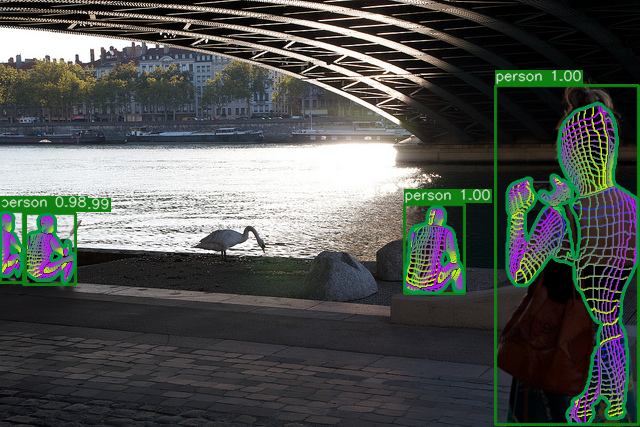}
		\includegraphics[width=4cm,height=3cm]{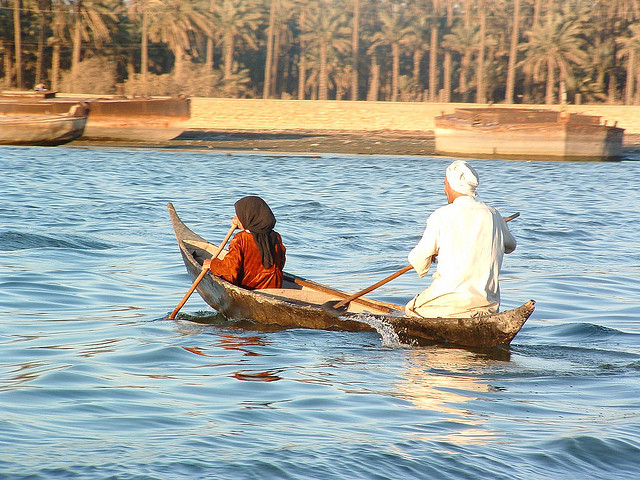}
		\includegraphics[width=4cm,height=3cm]{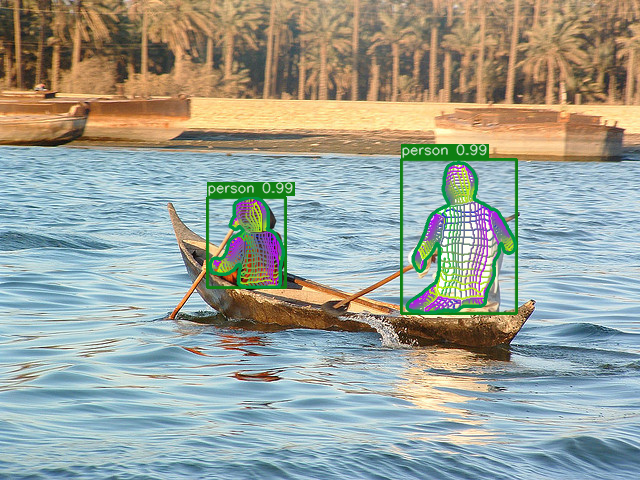}
	\end{minipage}
	\centering
	\captionsetup{font={normalsize}, justification=raggedright}
	\caption{
		Qualitative results of UV R-CNN. Left: original image of DensePose-COCO validation subset, Right: visualization of UV R-CNN predictions.}
	\label{fig:pic_static_info}
\end{figure*}

We also progressively adopt the DP-Loss, BWS, and the already proven practical improvements of dense pose network to boost the performance of our final model. Serialization experiments with ResNet-50-FPN on DensePose-COCO dataset are represented in Table~\ref{table:r50ablation}, our final model achieves 65.0\% $AP_{gps}$ and 66.1\% $AP_{gpsm}$.

\subsection{Comparisons with State-of-the-Art Methods}
\label{sec:SOTA}
In order to further demonstrate the effectiveness, we compare the proposed UV R-CNN to the state-of-the-art methods on DensePose-COCO dataset. We adopt ResNet-50-FPN as backbone, respectively, and corresponding results are shown in Table~\ref{table:SOTA}. Combined with all the proposed components, UV R-CNN achieves 65.0 $APgps$, which yields 14.0 improvements to DensePose-RCNN and 6.7 improvements to Parsing R-CNN. Note that state-of-the-art method KTN\cite{ktn} uses external knowledge learned from keypoint annotations of the MSCOCO dataset, and its best competitor AMA-Net\cite{amanet} uses human body mask and human part mask prediction to supervise the learning of U/V coordinates externally. Without auxiliary supervision and external knowledge, the proposed UV R-CNN still achieves competitive results.

\subsection{Qualitative Results}
\label{sec:visualization}

In this subsection, we provide additional qualitative results to further demonstrate the performance of  UV R-CNN. In Fig.~\ref{fig:pic_static_info}, we show qualitative results generated by our method.

\label{sec:experiments}

\section{Conclusion}
In this work, we analyse the loss formulation and optimizing strategy for deep learning dense pose estimation models, and propose a stable and efficient instance-level dense pose estimation architecture, named UV R-CNN, equiped with a novel human surface point regression loss function and a balanced loss weighting strategy. Without external knowledge and auxiliary supervision from other human instances analyzing tasks and data sources, the proposed UV R-CNN achieves competitive results with the state-of-the-art methods on DensePose-COCO benchmark. Our approach explores the fundamental problems of optimizing dense human pose estimator and verifies the effectiveness of our loss modification and optimizing strategy to solve these problems.



\bibliographystyle{elsarticle-num} 
\bibliography{ref}


%% \begin{thebibliography}{00}


%% \bibitem{}

%% \end{thebibliography}

\end{document}